# Reward function shape exploration in adversarial imitation learning: an empirical study

Yawei Wang
The Shenzhen International Graduate School
Tsinghua University
Shenzhen, China
e-mail: wyw18@mails.tsinghua.edu.cn

Xiu Li[*]
The Shenzhen International Graduate School
Tsinghua University
Shenzhen, China
e-mail: li.xiu@sz.tsinghua.edu.cn

*Abstract*—For adversarial imitation learning algorithms (AILs), no true rewards are obtained from the environment for learning the strategy. However, the pseudo rewards based on the output of the discriminator are still required. Given the implicit reward bias problem in AILs, we design several representative reward function shapes and compare their performances by large-scale experiments. To ensure our results' reliability, we conduct the experiments on a series of Mujoco and Box2D continuous control tasks based on four different AILs. Besides, we also compare the performance of various reward function shapes using varying numbers of expert trajectories. The empirical results reveal that the *positive logarithmic reward function* works well in typical continuous control tasks. In contrast, the so-called unbiased reward function is limited to specific kinds of tasks. Furthermore, several designed reward functions perform excellently in these environments as well.

## I. INTRODUCTION

Deep reinforcement learning (DRL) has achieved extraordinary performance in Chess and cards [1, 2], computer games [3-5], and robot manipulation [6] in recent years. On most of these domains, it is evident for humans to design rewards, which reflect what agent's behavior is favorable. However, designing reasonable rewards by hand may be difficult and time-consuming when it comes to real-world tasks. Imitation learning (IL) [7] comes in such situations, which learns directly from expert demonstrations without reward signals from the environment. Generally, we can divide imitation learning into two categories: behavioral cloning (BC) and inverse reinforcement learning (IRL).

BC [8, 9] is the simplest method for IL, which considers imitation learning a supervised learning problem. Without interacting with the environment during training, BC is the first choice for IL when enough expert demonstration is available. However, BC often fails to imitate expert policy in real-world scenes due to the *compounding error* problem [10]. In these cases, it is hard for humans to obtain ample specialist demonstrations.

IRL [11, 12] focuses on finding a reward function that could match the demonstrated behavior well, thus overcoming the *compounding error* problem. This reward function is then used to guide the agent for learning an imitation policy using standard reinforcement learning (RL) algorithms. However, the function search process is ill-posed because the expert data may correspond to multiple reward functions [13]. Besides, IRL suffers from tremendous computation due to the interplay between reward searching and RL optimization.

Generative Adversarial Imitation Learning (GAIL) [14] has recently become a popular model-free imitation learning framework because it outperforms BC and IRL in a variety of continuous control tasks. The training process of GAIL is similar to Generative Adversarial Nets (GANs) [15]. In GAIL, the discriminator is trained to distinguish whether state-action pairs come from the expert demonstrations or the interaction with the environment. The generator, i.e., the imitation policy, needs to generate interaction data to fool the discriminator as much as possible. While no true reward is available in GAIL, the pseudo reward related to the discriminator's output is still required to guide the imitation policy's learning. However, the commonly used logarithmic reward functions may bias to different tasks.

The IL community has been paying attention to unbiased reward function for robustly learning in various tasks. Researchers proposed the neutral reward function to deal with the variability [16] and the multiple terminal states [17] in the environment. The Discriminator-Actor-Critic (DAC) method [18] explicitly handles absorbing state transitions by learning the reward associated with these states, mitigating the survival bias incurred by the *positive logarithmic reward function*. However, experimental results in these works may be one-sided and a little unconvincing. The proposed methods can handle some particular challenges but fail to imitate well in typical continuous control environments.

Although IL has gotten rid of reward engineering, reward function shapes still play an essential role in adversarial imitation learning algorithms (AILs). In this paper, we enlarge the reward function pool by introducing other representative reward function shapes. Based on this reward function pool, we combine various GANs and RL algorithms to form four AILs and examine them in popular Mujoco [19] and Box2D tasks. We hope to provide a novel perspective to enhance people's perceptual knowledge for different reward functions' performance in typical continuous control tasks. Meanwhile, we also try to answer the following specific questions: 1) How the traditional positive and negative logarithmic reward functions behave in these tasks on earth? 2) Is the so-called unbiased, neutral reward function practical for typical continuous control tasks? 3) What are the results if we perform imitation learning using non-logarithmic reward functions? Our empirical observations suggest that the *positive logarithmic reward function* behaves surprisingly well in these tasks, conflicting with the results in [18]. We also notice

* The corresponding author.
This work was supported by the National Key R & D Program of China (No. 2020AAA0108303) and NSFC 41876098.

that the so-called unbiased, neutral reward function may not be as preferable as in specific environments. Besides, several other reward functions can also achieve excellent performance in these environments, which brings us a huge surprise. This paper is the first work to evaluate the reward function shapes in AILs on a large scale to the best of our knowledge.

## II. BACKGROUND

### A. Markov Decision Process

We consider the finite-horizon Markov decision process (MDP) defined by the tuple $(S, A, p, r)$. When learning policy in continuous action space, the state space $S$ and the action space $A$ are continuous. The state transition probability $p$ represents the probability of the following state $s_{t+1}$ given the current state $s_t$ and action $a_t$. When executing an action $a_t$, the environment will emit a bounded reward $r_t \in [r_{min}, r_{max}]$ defined by the human expert. An agent's purpose is to learn a policy for maximizing the cumulative discounted rewards from the collected transitions:

$$\mathbb{E}_\tau \sum_{t=0}^{T} \gamma^t r(s_t, a_t), \quad (1)$$

in which $\gamma$ is the discount factor and $\tau = \{s_1, a_1, \ldots, a_{T-1}, s_T\}$ denotes the whole interaction trajectory.

### B. Generative Adversarial Imitation Learning

We take GAIL [14] as the base algorithm, which combines the inspiration of maximum entropy IRL [13] and Generative Adversarial Networks (GANs) [15]. GAIL trains a binary classifier called the discriminator to differentiate between transitions sampled from the expert and those generated by the imitation policy. Unlike in standard GANs, the generator in GAIL is provided a reward for confusing the discriminator. The reward is positively correlated with the probability that the generated state-action pairs come from an expert trajectory, which is then maximized via on-policy RL optimization algorithms, such as trust region policy optimization (TRPO) [20] or proximal policy optimization (PPO) [21]. Specifically, our goal is to find a saddle point $(\pi, D)$ of the following expression:

$$\mathbb{E}_\pi[\log(D(s, a))] + \mathbb{E}_{\pi_E}[\log(1 - D(s, a))]. \quad (2)$$

To find the saddle point, we first need to introduce function approximation for $\pi$ and $D$. We parameterize the policy network $\pi_\theta$ and the discriminator network $D_w: S \times A \to (0,1)$ with weights $\theta$ and $w$, respectively. Then, the alternating optimization process between the imitation policy and the discriminator is carried out. Specifically, we alternate between a stochastic gradient step on $w$ to decrease (2) concerning $D$ and an RL step on $\theta$ to increase (2) concerning $\pi$. The RL step serves the same effect as it does in conventional reinforcement learning algorithms: aiming to maximize the cumulative expected reward.

The discriminator can be considered as a local cost function offering learning reward signals to the imitation policy. That is to say, taking a policy step that decreases the expected cost for the cost function $c(s, a) = -\log D_\pi(s, a)$ will naturally increase the cumulative expected reward for the reward function $r(s, a) = \log D_\pi(s, a) \in [-\infty, 0]$, i.e., the *negative logarithmic reward function*. The other frequently-used reward function $r(s, a) = -\log(1 - D_\pi(s, a)) \in [0, +\infty]$ correlates to the probability $D_\pi(s, a)$ positively, referred to as the *positive logarithmic reward function*. These two classical logarithmic reward functions are the first shapes utilized in GAIL.

### C. Wasserstein Generative Adversarial Imitation Learning

In GAIL, the discriminator's optimal loss function is the Jensen-Shannon (JS) divergence. The RL algorithm aims to minimize the JS divergence for the imitator and the expert's occupancy measure. However, the vanishing gradients problem caused by the JS divergence may harm the adversarial learning process [22]. In [22, 23], the Wasserstein GANs proposed using the Earth-Mover distance, i.e., the Wasserstein distance, to measure the divergence between two distributions: smooth and continuous everywhere. In other words, the gradients of weights $w$ are always available and meaningful.

Consequently, this improvement for GAN is borrowed to form other AILs [24], and we uniformly call them Wasserstein Generative Adversarial Imitation Learning algorithms (WGAILs), the objective function of which is as follows:

$$min_\pi max_D [\mathbb{E}_\pi(D(s, a)) - \mathbb{E}_{\pi_E}(D(s, a))], \quad (3)$$

where the Lipschitz condition should constrain the weights $w$. In the original WGAN paper [22], the weights are clamped to a fixed box after each gradient update. One improved training method of WGAN is proposed in [23]: a penalty is enforced to the discriminator's norm concerning its input.

## III. RELATED WORK

AILs [14, 24] bypass the exact reward functions, trying to directly recover the expert policy with the discriminator's generated reward. AILs have an outstanding advantage in sample efficiency in terms of expert data. Nevertheless, the generative reward may be unstable and does not guarantee expert trajectories' optimality.

### A. Learning Robust Rewards with AIRL

The AIRL method [16] resembles GAIL and GAN-GCL [25] regarding the generator and the discriminator's alternate training procedure. However, in this paper, a unique structure is placed to form a restricted discriminator. Depending on this discriminator, we can recover the reward to a constant if the ground truth reward is only a state function. The reward update formula is as follows:

$$r(s_t, a_t) \leftarrow \log(D(s_t, a_t)) - \log(1 - D(s_t, a_t)). \quad (4)$$

This form of reward function combines the positive and negative logarithmic reward functions. We refer to it as the *combination reward function*. Rewards learned with AIRL can transfer effectively under variation in the underlying tasks, in contrast to unmodified IRL methods, which tend to recover brittle rewards that do not generalize well, and GAIL, which does not recover reward functions at all.

### B. Discriminator Actor Critic

GAIL may suffer from different biases induced by the classical logarithmic reward functions. In [18], the authors propose that these biases can account for one reason – the terminal state's reward is implicitly set to zero in all the situations. So, they define a set of *absorbing states* $s_a$ [26] that the agent enters after the end of an episode, gaining zero rewards and always transiting to itself.

To address the implicit biases in GAIL, DAC learns an additional reward item apart from the discriminator reward for the *absorbing states* in the expert demonstrations and the interaction trajectories. We consider a trajectory of T time steps, of which the return for the final state $s_T$ is defined as follows:

$$R_T(s,a) = r(s_T, a_T) + \sum_{t=T+1}^{\infty} \gamma^{t-T} r(s_a, \cdot), \quad (5)$$

with a learned reward function $r(s_a, \cdot)$ in DAC instead of just $R_T(s,a) = r(s_T, a_T)$ in standard GAIL setting. This extra learned reward for the terminal state eliminates the bias towards avoiding the *absorbing states* in survival-based tasks and transitioning to the *absorbing states* in goal-based tasks. With this formulation, the state occupancy of the expert and the agent's terminal state is also considered matching, hence alleviating the induced biases. However, DAC modifies a finite-horizon environment into an infinite-horizon one, and the wrappers are demanded to augment the environments with additional terminal states.

### C. Neutral Reward Functions

Reference [17] proposes the neutral reward function to denote real-valued reward functions. By coincidence, the reward function used in [17] is $r(s,a) = \log(D(s,a)) - \log(1 - D(s,a))$ as well, the same as in [18]. It is worth noticing that the *combination reward function* is equal to the discriminator's actual output, which has the range $(-\infty, +\infty)$ and is centrosymmetric. Consider an oracle discriminator that gives positive reward $R$ when the agent takes the expert action, and negative reward $-R$ otherwise. Under the circumstances, the agent will always get fewer rewards if it chooses to loop in the environment or follow shorter non-expert trajectories. So, the agent will learn to imitate the expert favorably and overcome all the biases.

## IV. METHOD

### A. Tasks with Different Characteristics

Survival-based tasks: the agent will receive per-step survival rewards from the environment, which are always positive. However, if the agent incautiously reaches a "bad state", the episode will terminate soon. Therefore, the agent is encouraged to stay in a set of "good states" to maximize its return.

Goal-based tasks: the agent can get a considerable reward when reaching a specific state in the environment. Moreover, additional bonuses may be given for awarding the completion of "subtasks" along the way to the final goal state. Meanwhile, the agent is also penalized for merely staying in the environment instead. To maximize the return with the positive success reward and the negative per-step penalty, the agent must complete the task as soon as possible.

Other more complex tasks: there is the goal state that the agent should learn to reach and other terminal states that lead to the end of an episode because of the agent's death. For goal-based environments where multiple termination conditions exist, excessive penalization may bias the agent to the shorter path to end the episode but not finish the task itself.

### B. Reward Bias in GAIL

As we have introduced in II.B, two kinds of reward function shapes are commonly used in GAIL and its following methods [14, 24, 27] to guide the agent for imitating the expert, i.e., the *positive logarithmic reward function* and the *negative logarithmic reward function*. However, both the positive and negative reward functions can induce implicit biases, which may hinder imitating. Next, we will make a more in-depth exploration of these two forms of reward functions, ad analyze the reasons for the introduction of biases.

The *positive logarithmic reward function* may suffer from survival bias. As the recovered reward will never be negative, it cannot reflect the true reward in goal-based tasks where the agent must solve the task as quickly as possible. Consider an ideal situation that the discriminator is perfect and provides a positive reward if the expert action is performed, and zero otherwise. In such a case, as long as the discount factor $\gamma \geq 0.618$, the agent is promised to gain more rewards supposing looping in the environment than following the expert trajectories [17]. The positive reward incentivizes the agent to move into loops or taking small actions for permanently closing to the states in the expert trajectories. Therefore, using this reward function will introduce survival bias and result in sub-optimal solutions in goal-based tasks.

The *negative logarithmic reward function* may lead to penalty bias. If we adopt the negative reward variant, the agent is penalized every step for forcing the episode to finish earlier. This negative reward may be beneficial for goal-based tasks in which the fastest way to complete the task is preferred. However, this negative reward function may not be applicable in the following two cases: 1) if the task is survival-based, the agent needs to explore more in the early stage of training. The per-step penalty inevitably impedes the agent to learn to stay in the "good" states. 2) if there are faster ways to only break off an episode in the goal-based tasks, the agent will greedily pursue that without completing the task, which deviates from the original goal.

### C. Representative Reward Function Shapes

We design other five shapes of reward functions in addition to the original three logarithmic reward functions. We demonstrate the eight different reward function shapes in fig. 1. And the details of these representative reward functions are as followed.

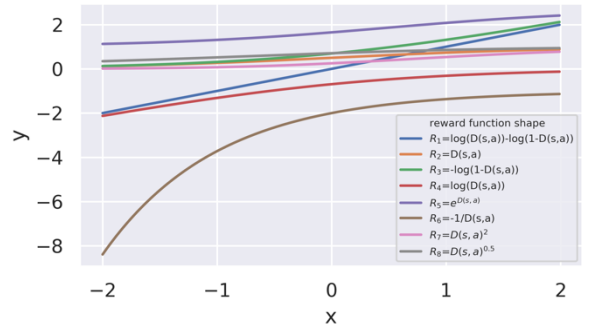

Fig. 1. Different shapes of reward functions.

*1)* $R_1 = \log(D(s,a)) - \log(1 - D(s,a)) = x$ : this is the *combination reward function*, which is linear and has a value range $[-\infty, +\infty]$.

*2)* $R_2 = D(s,a) = sigmoid(x)$: this is a new nonlinear positive reward function, and its value range is $[0,1]$.

*3)* $R_3 = -\log(1 - D(s,a))$: this shape is the *positive logarithmic reward function* whose value range is $[0, +\infty]$.

*4)* $R_4 = \log(D(s,a))$: this is the *negative logarithmic reward function*, which has a value range $[-\infty, 0]$.

*5)* $R_5 = e^{D(s,a)}$ : this is the exponential shape of the reward function and has a value range $[1, e]$.

*6)* $R_6 = -1/D(s,a)$ : this is the inverse proportional reward function and has a value range $[-\infty, -1]$.

*7)* $R_7 = D(s,a)^2$: this type of reward shape is a convex power function whose value range is $[0,1]$.

*8)* $R_8 = \sqrt{D(s,a)}$: this is another kind of power function shape reward and has the same value range as $R_2$ and $R_7$, but it is nonconvex.

## V. EXPERIMENTS

In this section, we will perform a series of experiments for the typical continuous control tasks of Mujoco and Box2D to answer the following questions:

- How the traditional positive and negative logarithmic reward functions perform in these tasks?
- Is the unbiased, neutral reward function superior in all cases?
- What are the results if other shapes of reward functions are applied?

### A. Tasks, Experts, and Algorithms

The tasks we test are five Mujoco environments: Ant-v2, HalfCheetah-v2, Hopper-v2, Reacher-v2, and Walker2d-v2 and a Box2D task BipedalWalker-v3. Unlike the particular tasks in [17], the tasks we used are all typical continuous control tasks. We perform experiments in these environments for evaluating different reward functions, not for dealing with specific situations. Since the Twin Delayed Deep Deterministic policy gradient algorithm (TD3) [28] has achieved state-of-the-art performance in the tasks of Mujoco, we choose to generate expert demonstrations using trained TD3 agents. For each of our test tasks, the expert data contain 20 expert trajectories.

To make sure that the empirical results are reliable and trustworthy, we use two RL algorithms, PPO and TD3, to learn the policy and two GANs, i.e., GAN and WGAN, to train the discriminator, which results in four AILs: GAIL-PPO, GAIL-TD3, WGAIL-PPO, and WGAIL-TD3. We use the code provided by Kostrikov et al. [29] to run GAIL-PPO and form our WGAIL-PPO algorithm. Moreover, we implement the GAIL-TD3 and WGAIL-TD3 algorithms by Pytorch based on the official TD3 code. For GAIL-PPO and WGAIL-PPO, the actor and critic structure are a two-layer MLP with Tanh activations and 64 hidden units. For GAIL-TD3 and WGAIL-TD3, we use the same architecture as in [28]: a two-layer MLP with Relu activations and 256 hidden units. We also use a two-layer MLP with Tanh activations and 100 hidden units for the discriminator in all four methods. We train all the networks with the Adam optimizer and the learning rate 3e-4. To make the training procedure more stable, we regularize the discriminator using gradients penalty [23]. We perform evaluating every 2000 steps and compute average returns on five evaluations, and five trials with different random seeds are conducted for each experiment. Furthermore, we have normalized the returns to [0,1] for more clear comparison.

### B. Results for Various Reward Functions

We want to know how the reward function shapes influences adversarial imitation learning for different tasks. Based on the reward functions in IV.C, we conduct GAIL-PPO experiments for various continuous control tasks, using five expert trajectories. The results of GAIL-PPO for eight reward functions are shown in TABLE I.

We can find that the *positive logarithmic reward function* $R_3$ behaves universally well in all six environments. In contrast, the *negative logarithmic reward function* $R_4$ can only effectively imitate the expert in the HalfCheetah-v2 and Reacher-v2 environments. However, the *combination reward function* $R_1$, which has shown its superiority on particular domains [17], is not suitable for Ant-v2, Hopper-v2, and Walker2d-v2 and performs inconsistently for different trials in BipedalWalker-v3.

Among the other five designed reward functions, the positive reward functions $R_2$, $R_7$, and $R_8$ can gain the same excellent performance as $R_3$ for all the tasks. However, the positive reward function $R_5$ fails to imitate in Ant-v2, and the learned policies are not good in BipedalWalker-v3, Hopper-v2, and Walker2d-v2. We speculate that the value range [1, e] damages the learning in the early training procedure because a reward no less than one is given even the generated transitions are not likely real, which sometimes results in an inferior imitation policy. The designed negative reward function $R_6$ performs similarly to the reward function $R_1$.

TABLE I. NORMALIZED RETURN OF DIFFERENT REWARD FUNCTIONS FOR GAIL-PPO

|  | Ant-v2 | BipedalWalker-v3 | HalfCheetah-v2 | Hopper-v2 | Reacher-v2 | Walker2d-v2 |
| --- | --- | --- | --- | --- | --- | --- |
| $R_1$ | 0.014 | 0.324 | 1.105 | -0.004 | 0.912 | -0.001 |
| $R_2$ | 1.106 | 0.895 | 1.099 | 1.007 | 0.924 | 1.080 |
| $R_3$ | 1.101 | 0.966 | 1.101 | 1.007 | 0.913 | 1.069 |
| $R_4$ | 0.014 | 0.020 | 1.104 | -0.004 | 0.887 | -0.001 |
| $R_5$ | -0.008 | 0.698 | 1.107 | 0.562 | 0.913 | 0.243 |
| $R_6$ | 0.017 | 0.398 | 1.085 | 0.008 | 0.906 | 0.014 |
| $R_7$ | 1.106 | 0.944 | 1.097 | 1.006 | 0.921 | 1.072 |
| $R_8$ | 1.077 | 0.959 | 1.102 | 1.005 | 0.910 | 1.017 |

*C. Results for Varying Expert Trajectories*

In GAIL and its following methods, researchers often compare the results of different approaches when the size of the used expert trajectories varies, manifesting the sample efficiency of AILs in terms of expert data. We further choose to use 1, 10, and 20 trajectories to evaluate the performance of GAIL-PPO in different environments and summarize the results in Fig. 2.

The four positive reward functions $R_2$, $R_3$, $R_7$, and $R_8$ can achieve nearly consistent performance for all six tasks. In Ant-v2, these rewards lead to a suboptimal imitation policy when only one expert trajectory is provided. In Reacher-v2, there is an evident tendency that the more the given expert trajectories are, the better these positive reward functions will behave. Another positive reward function $R_5$ can achieve excellent performance in HalfCheetah-v2 and Reacher-v2. However, in BipedalWalker-v3, Hopper-v2, and Walker2d-v2, the agent trained using $R_5$ behaves relatively poorly and has undulating imitation results influenced by the number of expert data. And $R_5$ fails to imitate the expert for the task of Ant-v2, even if 20 expert trajectories are used.

The *combination reward function $R_1$* is merely successful in HalfCheetah-v2 when the expert trajectories vary. In the task of Reacher-v2, the rate of successful imitation decreases to less than 60% if only one expert trajectory is supplied. In BipedalWalkerv3, the performance has huge variance using the reward function $R_1$. As for the other three tasks, the reward function $R_1$ does not help imitate the expert strategy.

The *negative logarithmic reward function $R_4$* is effective in HalfCheetah-v2 and Reacher-v2, but its performance in Reacher-v2 declines to a 50% successful rate if one expert trajectory is provided. In BipedalWalker-v3, where other reward functions succeed in imitating to a certain extent, $R_4$ can imitate nothing. The designed negative reward function $R_6$ has a similar imitation performance compared to $R_1$ in all tasks.

*D. Results for Different AILs*

To ensure that the above empirical conclusions are not valid only for GAIL-PPO, we have done parallel experiments for WGAIL-PPO, GAIL-TD3, and WGAIL-TD3 using 1, 5, 10, and 20 expert trajectories. Due to the space limitations, we simply demonstrate the results for different reward functions in BipedalWalker-v3, as shown in Fig. 3. In these experiments, we use five trajectories of expert data for each task. We have conducted complete experiments in all the environments, but the consequences of BipedalWalker-v3 are representative enough to deliver the performance for most reward functions.

We can discover from Fig. 3 that all four AILs can imitate the expert well using positive reward functions $R_2$, $R_3$, $R_7$, and $R_8$. These results imply that one reward function may lead to a similar imitation policy using different AILs. The positive reward function $R_5$ can achieve slightly worse performance compared to the other four positive reward functions.

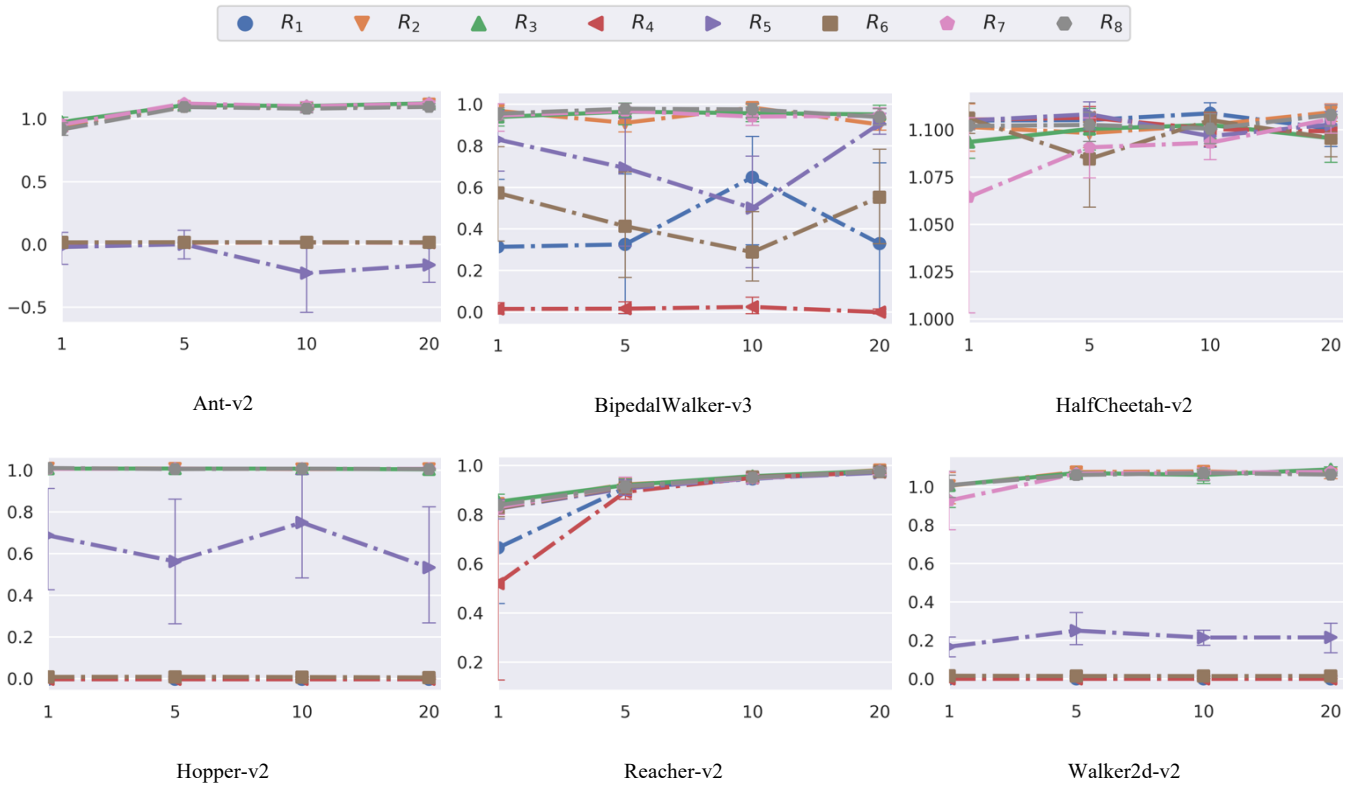

Fig. 2. Performances on different reward functions vary when the number of expert trajectories changes. The X-axis represents the number of used expert trajectories, and the Y-axis represents the normalized episodic return.

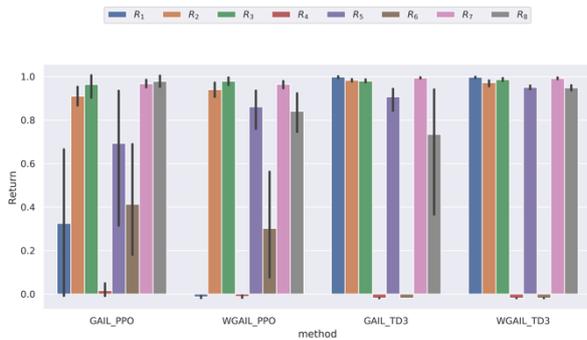

Fig. 3. Performances on different reward functions vary in BipedalWalker using various AILs.

If the reward function $R_6$ is used, GAIL-PPO and WGAIL-PPO can succeed in imitating to a certain extent, but GAIL-TD3 and WGAIL-TD3 fail. However, if we use the reward function $R_1$, GAIL-TD3 and WGAIL-TD3 achieve extraordinary imitation success rates, but GAIL-PPO and WGAIL-PPO perform poorly. All the methods cannot recover the expert policy if the *negative logarithmic reward function* $R_4$ guides the imitator.

## VI. CONCLUSION AND FUTURE WORK

There are a series of empirical conclusions we can figure out from the results in V.B, V.C, and V.D.

1. The *positive logarithmic reward function* $R_3$ is good enough for typical continuous control tasks of Mujoco and Box2D. Maybe the survival bias is not widespread as imaged. In contrast, the penalty bias is much more common because the *negative logarithmic reward function* $R_4$ often fails.

2. The so-called unbiased, neutral reward function $R_1$ performs poorly in some environments, though it could achieve excellent performance in other cases.

3. Several newly designed reward function shapes are competent for imitating the expert strategy in complex continuous control tasks.

4. The factors that influence how a reward function works may be the environments, the number of the expert trajectories, the used AIL method, even a random seed for running the experiment.

So far, many of the reward function shapes are partially applicable, and the more general reward shaping methods for AILs are highly wanted.

REFERENCES

[1] D. Silver et al., "A general reinforcement learning algorithm that masters chess, shogi, and Go through self-play," Science, vol. 362, no. 6419, pp. 1140-1144, 2018.
[2] N. Brown and T. Sandholm, "Superhuman AI for multiplayer poker," Science, vol. 365, no. 6456, pp. 885-890, 2019.
[3] V. Mnih et al., "Human-level control through deep reinforcement learning," Nature, vol. 518, no. 7540, pp. 529-533, 2015.
[4] O. Vinyals et al., "Grandmaster level in StarCraft II using multi-agent reinforcement learning," Nature, vol. 575, no. 7782, pp. 350-354, 2019.
[5] D. Ye et al., "Towards Playing Full MOBA Games with Deep Reinforcement Learning," in Advances in Neural Information Processing Systems, 2020, pp. 621-632.
[6] S. Levine, P. Pastor, A. Krizhevsky, J. Ibarz, and D. Quillen, "Learning hand-eye coordination for robotic grasping with deep learning and large-scale data collection," The International Journal of Robotics Research, vol. 37, no. 4-5, pp. 421-436, 2017.
[7] A. Hussein, M. Gaber, E. Elyan, and C. Jayne, "Imitation Learning: A survey of learning methods," ACM Computing Surveys, vol. 50, no. 2, pp. 1-35, 2017.
[8] M. Bain and C. Sammut, "A Framework for Behavioural Cloning," in Machine Intelligence 15, 1995, pp. 103--129.
[9] S. Ross, G. Gordon, and D. Bagnell, "A reduction of imitation learning and structured prediction to no-regret online learning," in Proceedings of the fourteenth international conference on artificial intelligence and statistics, 2011, pp. 627--635.
[10] S. Ross and D. Bagnell, "Efficient reductions for imitation learning," in Proceedings of the thirteenth international conference on artificial intelligence and statistics, 2010, pp. 661--668.
[11] A. Ng and S. Russell, "Algorithms for inverse reinforcement learning," in Proceedings of the Seventeenth International Conference on Machine Learning, 2000, pp. 663–670.
[12] P. Abbeel and A. Ng, "Apprenticeship learning via inverse reinforcement learning," in Proceedings of the twenty-first international conference on Machine learning, 2004, p. 1.
[13] B. Ziebart, A. Maas, J. Bagnell, and A. Dey, "Maximum entropy inverse reinforcement learning," in Proceedings of the 23rd national conference on Artificial intelligence, 2008, pp. 1433--1438.
[14] J. Ho and S. Ermon, "Generative adversarial imitation learning," in Proceedings of the 30th International Conference on Neural Information Processing Systems, 2016, pp. 4572--4580.
[15] I. Goodfellow et al., "Generative adversarial nets," in Proceedings of the 27th International Conference on Neural Information Processing Systems, 2014, pp. 2672–2680.
[16] J. Fu, K. Luo, and S. Levine, "Learning Robust Rewards with Adversarial Inverse Reinforcement Learning," in International Conference on Learning Representations, 2018.
[17] R. Jena, S. Agrawal, and K. Sycara, "Addressing reward bias in Adversarial Imitation Learning with neutral reward functions," unpublished.
[18] I. Kostrikov, K. Agrawal, D. Dwibedi, S. Levine, and J. Tompson, "Discriminator-Actor-Critic: Addressing Sample Inefficiency and Reward Bias in Adversarial Imitation Learning," in International Conference on Learning Representations, 2018.
[19] E. Todorov, T. Erez, and Y. Tassa, "Mujoco: A physics engine for model-based control," in 2012 IEEE/RSJ International Conference on Intelligent Robots and Systems, 2012, pp. 5026--5033.
[20] J. Schulman, S. Levine, P. Abbeel, M. Jordan, and P. Moritz, "Trust region policy optimization," in Proceedings of the 32nd International Conference on Machine Learning, 2015, pp. 1889--1897.
[21] J. Schulman, F. Wolski, P. Dhariwal, A. Radford, and O. Klimov, "Proximal policy optimization algorithms," unpublished.
[22] M. Arjovsky, S. Chintala, and L. Bottou, "Wasserstein generative adversarial networks," in International conference on machine learning, 2017, pp. 214--223.
[23] I. Gulrajani, F. Ahmed, M. Arjovsky, V. Dumoulin, and A. Courville, "Improved training of wasserstein GANs," in Proceedings of the 31st International Conference on Neural Information Processing Systems, 2017, pp. 5769--5779.
[24] M. Zhang et al., "Wasserstein Distance guided Adversarial Imitation Learning with Reward Shape Exploration," in 2020 IEEE 9th Data Driven Control and Learning Systems Conference (DDCLS), 2020, pp. 1165--1170.
[25] C. Finn, P. Christiano, P. Abbeel, and S. Levine, "A connection between generative adversarial networks, inverse reinforcement learning, and energy-based models," unpublished.
[26] R. Sutton and A. Barto, Reinforcement learning: An introduction, MIT press, 2018.
[27] F. Sasaki, T. Yohiraa, and A. Kawaguchi, "Sample efficient imitation learning for continuous control," in International Conference on Learning Representations, 2018.
[28] S. Fujimoto, H. Hoof, and D. Meger, "Addressing Function Approximation Error in Actor-Critic Methods," in International Conference on Machine Learning, 2018, pp. 1587--1596.
[29] I. Kostrikov, "PyTorch Implementations of Reinforcement Learning Algorithms," unpublished.